\documentclass{article}
\usepackage[margin=1in]{geometry}

\usepackage{graphicx}
\usepackage{booktabs}
\usepackage{multirow}
\usepackage{wrapfig}

\usepackage{amsthm}
\usepackage{amssymb}

\newtheorem{theorem}{Theorem}
\newtheorem{proposition}{Proposition}
\newtheorem{remark}{Remark}

\usepackage{hyperref}

\usepackage{amsmath}
\DeclareMathOperator*{\argmin}{arg\,min}  % in preamble
\usepackage{cleveref}

\begin{document}

% ---------------------------------------------------------------
% TODO REVIEW: Replace with your title
\title{ODE-free Neural Flow Matching for One-Step Generative Modeling} 

\author{Xiao Shou\\
Baylor University}
\date{}

\maketitle

\begin{abstract}
Diffusion and flow matching models generate samples by learning
time-dependent vector fields whose integration transports noise to data,
requiring tens to hundreds of network evaluations at inference.
We instead learn the transport map directly.
We propose Optimal Transport Neural Flow Matching (OT-NFM), an
ODE-free generative framework that parameterizes the flow map with neural flows, enabling true one-step generation with a single forward pass.
We show that naive flow-map training suffers from \emph{mean collapse},
where inconsistent noise–data pairings drive all outputs toward the data
mean. We prove that consistent coupling is necessary for non-degenerate
learning and address this using optimal transport pairings with scalable
minibatch and online coupling strategies.
Experiments on synthetic benchmarks and image generation tasks
(MNIST and CIFAR-10) demonstrate competitive sample quality while
reducing inference to a single network evaluation.
  \textbf{Keywords:} Neural Flow $\cdot$ Flow Matching $\cdot$ Optimal Transport
\end{abstract}

\section{Introduction}

Generative modeling has undergone a remarkable transformation in recent years.
Diffusion models~\cite{sohldickstein2015deep, ho2020denoising, song21score}
demonstrated that high-fidelity synthesis can be achieved by learning to reverse
a gradual noising process.
Flow matching~\cite{lipman2023flow, tong2024improving}
reformulated this paradigm in a more direct deterministic framework by learning a
time-dependent velocity field whose integration transports noise to data. Despite their success, most diffusion and flow-matching methods learn time-dependent local vector fields, scores or velocities, whose numerical integration defines the global transport map. Consequently, sample generation requires iterative ODE/SDE solvers, typically involving tens of neural network evaluations.

One-step generation has consequently become a central challenge.
Consistency models~\cite{song2023consistency, song2024improved} impose
self-consistency constraints along learned trajectories to collapse sampling to
a single step, but require careful discretization curricula and can be unstable
to train.
%Distillation-based methods~\cite{liu2023flow} transfer the quality of a many-step teacher into a one-step student, introducing a costly two-stage pipeline.
MeanFlow~\cite{geng2025mean} reformulates the problem through average velocity
fields and achieves impressive one-step results from scratch, yet still
fundamentally trains a velocity-based model requiring ODE integration at
inference.
A natural alternative is to sidestep velocity fields entirely and learn the
\emph{flow map}, the direct mapping from initial 
noise to data at time $t$, so that generation reduces to a single forward pass with no ODE solver. 

% In this work, we propose \textbf{Optimal Transport Neural Flow Matching}
% (OT-NFM), a framework that instantiates this idea via neural
% flows~\cite{bilovs2021neural}.
% We show that naive training of the flow map fails: without a structured
% coupling between noise and data, the squared-loss objective drives the model
% toward the global data mean --- a degenerate solution we term \emph{mean
% collapse}.
% This failure is unique to neural flows, which condition on the fixed initial
% noise $\mathbf{x}_0$ and have no mechanism to self-correct, unlike flow
% matching which conditions on the evolving state $\mathbf{x}_t$ at each
% integration step.
% Optimal transport resolves this by providing each $\mathbf{x}_0^{(i)}$ a
% geometrically consistent target throughout training, making OT \emph{necessary}
% rather than merely beneficial --- a strictly stronger role than it plays in
% flow matching. To this end, we summarize our contributions in the following:
% \begin{itemize}
%  \item We propose OT-NFM for ODE-free fast 1-step image generation and we extend the trajectories from various curveous ones. 
%   \item We identify the mean collapse problem for learning trajectory maps with neural flow models and OT as necessary condition for non-degenerate neural flow matching.
%   \item We implement two scalable OT strategies that achieves at $\mathcal{O}(B^3)$ cost per batch.  
%   \item Empirical validation on 2D distributions and image benchmarks 
%   demonstrating sharp one-step generation competitive with multi-step baselines.
% \end{itemize}
In this work, we propose \textbf{Optimal Transport Neural Flow Matching}
(OT-NFM), a framework that instantiates this idea via neural
flows~\cite{bilovs2021neural}, without distillation, consistency
constraints, or ODE solvers at inference.
We show that naive training of the neural flow map fails: without a structured
coupling between noise and data, the squared-loss objective drives the model
toward the global data mean --- a degenerate solution we term \emph{mean
collapse}.
This failure is unique to neural flows, which condition on the fixed initial
noise $\mathbf{x}_0$ and have no mechanism to self-correct, unlike flow
matching which conditions on the evolving state $\mathbf{x}_t$ at each
integration step.
Optimal transport resolves this by providing each $\mathbf{x}_0^{(i)}$ a
geometrically consistent target throughout training, making OT \emph{necessary}
rather than merely beneficial --- a strictly stronger role than it plays in
flow matching. Our contributions are as follows:

\begin{itemize}
  \item We propose OT-NFM, an ODE-free generative framework that learns the
  flow map directly via neural flows, enabling true one-step image generation
  with a single forward pass and no numerical integration at inference.
  We further analyze a family of interpolation trajectories and show that
  linear interpolation is the natural and most efficient choice for neural
  flow matching.

  \item We identify \emph{mean collapse}, a failure mode unique to neural
  flow models, and prove that an optimal transport coupling is a
  \emph{necessary} condition for non-degenerate flow map learning, a strictly
  stronger role than OT plays in conventional flow matching.

  \item We introduce two scalable OT coupling strategies: precomputed
  minibatch OT and online refinement via LOOM, both operating at
  $\mathcal{O}(B^3)$ cost per batch, making OT-NFM practical for large-scale
  training without precomputing the full $N{\times}N$ transport plan.

  \item We provide empirical validation on synthetic 2D benchmarks and image
  generation tasks (MNIST, CIFAR-10), demonstrating that OT-NFM achieves
  sharp, diverse one-step generation competitive with multi-step baselines at
  a fraction of the inference cost.
\end{itemize}

\section{Related Work}
\label{sec:related}

\paragraph{Diffusion and flow-based generative models.}
Diffusion probabilistic models~\cite{ho2020denoising,sohldickstein2015deep} learn to
reverse a fixed noisy diffusion process, achieving state-of-the-art sample
quality at the cost of hundreds of iterative denoising steps.
Score-based generative models~\cite{song21score} unify diffusion with
stochastic differential equations, enabling continuous-time formulations and
flexible noise schedules.
Flow matching~\cite{lipman2023flow, liu2023flow, albergo2023stochastic}
replaces stochastic dynamics with a deterministic velocity field governed by an ODE, enabling simulation-free training by
regressing conditional vector fields along interpolation paths between source
and target samples.
Optimal transport flow matching~\cite{tong2024improving, kornilov2024optimal}
incorporates OT couplings to straighten trajectories and reduce the number of
inference steps required. LOOM~\cite{davtyan2025faster} maintains an online OT coupling refined
iteratively during flow matching training, approximating the global transport plan without
precomputation. Our work departs from all of the above by learning the flow map directly rather than a velocity field, eliminating the need for
numerical ODE integration entirely via neural flow \cite{bilovs2021neural}. More recently, Curly Flow Matching~\cite{petroviccurly} extends flow
matching to non-gradient, periodic dynamics by constructing a Schr\"{o}dinger
bridge with a non-zero drift reference process, enabling trajectory inference
for systems such as single-cell RNA cycles that fundamentally cannot be
captured by gradient-field methods.

\paragraph{Few-step and one-step generation.}
Reducing inference cost from hundreds to few or one function evaluations has
been an active research direction.
Consistency models~\cite{song2023consistency, song2024improved} enforce
self-consistency along ODE trajectories, enabling one- or two-step sampling;
however, training can be unstable and requires a carefully designed
discretization curriculum.
Flow Map Matching~\cite{boffi2024flowmap} provides a principled mathematical
foundation for this family by directly parameterizing the two-time flow map of
an underlying dynamical model, unifying consistency models, consistency
trajectory models, and progressive distillation under a single framework.
Shortcut models~\cite{frans2025shortcut} introduce a self-consistency loss
over flow intervals, enabling one-step generation without distillation.
Inductive Moment Matching~\cite{zhou2025imm} models self-consistency of
stochastic interpolants at different time steps, achieving competitive one-step
results.
Rectified flow~\cite{liu2023flow} proposes iterative reflow to straighten
trajectories and accelerate sampling.
MeanFlow~\cite{geng2025mean} introduces average velocity fields and derives a
principled identity between average and instantaneous velocities, enabling
one-step generation from scratch without pre-training or distillation.
In contrast to all of these approaches, which either parameterize velocity
fields or rely on consistency constraints and ODE solvers at inference, our
method directly parameterizes the flow map via neural
flows~\cite{bilovs2021neural} and generates samples in a \emph{single forward
pass} without distillation, iterative refinement, or numerical integration.
Our work further distinguishes itself by grounding the flow map in an optimal
transport coupling, which we show is a \emph{necessary} condition for
non-degenerate training --- a role OT does not play in any of the
velocity-based methods above.

\section{Preliminary}

% \subsection{flow matching}
% Flow matching \cite{lipman2023} has emerged as a powerful framework for training continuous normalizing flows, but sampling requires solving an ODE, limiting inference speed. 

% We introduce a dual-path approach that learns both \textbf{V\_past} (displacement from start to current time) and \textbf{V\_future} (remaining displacement to endpoint), enabling one-step generation while preserving trajectory consistency.

We review two closely related classes of continuous-time generative models: \emph{flow matching} and \emph{neural flows}. While both aim to construct flow trajectory via continuous dynamics, they differ fundamentally in whether they parameterize the underlying velocity field or the resulting flow map. This distinction is central to the methodology developed in this work.

\subsection{Flow Matching}
\label{subsec:flow_matching}

Flow Matching~\cite{lipman2023flow} learns a continuous-time generative model
by regressing a velocity field that transports a base distribution $p_0$ to a
data distribution $p_1$.
Let $\mathbf{x}_t \in \mathbb{R}^d$ denote the state at time $t \in [0,1]$,
where $\mathbf{x}_0 \sim p_0$ (e.g.\ a standard Gaussian) and
$\mathbf{x}_1 \sim p_1$.
Given a coupling $\pi(x_0, x_1)$ and an interpolation scheme, the target
velocity along the path is
$
   \mathbf{u}(\mathbf{x}_t, t) = \frac{d\mathbf{x}_t}{dt}.
    \label{eq:target_vel}
$
A neural field $v_\theta(\mathbf{x}, t)$ is trained to match this target:
\begin{equation}
    \mathcal{L}_{\mathrm{FM}}(\theta) =
    \mathbb{E}_{(\mathbf{x}_0,\mathbf{x}_1)\sim\pi,\;t\sim\mathcal{U}(0,1)}
    \Bigl[\bigl\|v_\theta(\mathbf{x}_t, t) - \mathbf{u}(\mathbf{x}_t,t)
    \bigr\|^2\Bigr].
    \label{eq:fm_loss}
\end{equation}
Since the target velocity is defined analytically (e.g.\ via linear
interpolation), no numerical integration is needed during training.
At inference, samples are generated by integrating the learned field:
\begin{equation}
    \mathbf{x}_1 = \mathbf{x}_0 + \int_0^1 v_{\theta^*}(\mathbf{x}_s, s)\,ds,
    \qquad \mathbf{x}_0 \sim p_0,
    \label{eq:ode}
\end{equation}
which requires a numerical ODE solver, typically a Runge-Kutta method and
thus multiple network evaluations per sample.

\subsection{Neural Flows}
\label{subsec:neural_flows}

Neural flows~\cite{bilovs2021neural} are a class of continuous-time generative models that also describe the evolution of a state $\mathbf{x}_t \in \mathbb{R}^d$ over time, but differ fundamentally in how this evolution is parameterized.
Whereas flow matching \cite{lipman2023flow} (and Neural ODEs \cite{chen2018neural}) define dynamics implicitly via a velocity field, neural flows parameterize the \emph{flow map}—that is, the solution of the dynamics—directly.
Concretely, instead of specifying a differential equation for $\mathbf{x}_t$, neural flows model the state evolution explicitly as
\begin{equation}
\mathbf{x}_t = F_\theta(t, \mathbf{x}_0),
\end{equation}
where $F_\theta : [0,1] \times R^d \rightarrow R^d$ is a neural network that directly parameterizes the solution map from initial conditions and time to states.
To correspond to a valid continuous-time flow and probability path, the mapping $F_\theta$ is typically constructed to satisfy the following properties:
\begin{itemize}
    \item \textbf{Identity at the initial time.}
    The mapping recovers the identity at time zero, $F_\theta(0, \mathbf{x}_0) = \mathbf{x}_0$, ensuring that trajectories originate from their specified initial conditions.
    
    \item \textbf{Invertibility with respect to the state.}
    For any fixed $t$, the map $F_\theta(t, \cdot)$ is invertible, which guarantees uniqueness of trajectories and prevents intersections between solution curves corresponding to different initial states.
\end{itemize}
Under mild regularity conditions, these properties ensure the existence of an implicit time-dependent vector field whose integral curves are given by $F_\theta$, even though the vector field itself is never explicitly parameterized.

% We propose a neural flow for trajectory matching that directly parameterizes the transformation $F_\theta(t, \mathbf{x}) :=  \mathbf{x} + \int_0^t v_{\theta^*}(\mathbf{x}_s, s)ds $ mapping some sample state $\mathbf{x}$ from a base distribution to another sample state at time $t \in [0,1]$. Practically, we can map some noise $\mathbf{x}_0$ to a target sample $\mathbf{x}_1$. In contrast to (conditional) flow matching models which learn a velocity field, our approach learns the flow map itself. This formulation yields several key advantages: 1) stable training dynamics through direct trajectory supervision, 2) elimination of numerical integration during inference, 3) efficient one-step generation via a single evaluation $F_\theta(1, \mathbf{x}_0)$ 4) flexible trajectories to include linear and nonlinear ones. 
\section{Method}
We propose \emph{Neural Flow Matching} (NFM), a framework that directly
parameterizes the flow map $F_\theta : [0,1] \times \mathbb{R}^d \to
\mathbb{R}^d$ satisfying $F_\theta(0, \mathbf{x}_0) = \mathbf{x}_0$, mapping
a noise sample $\mathbf{x}_0 \sim p_0$ to a target sample
$F_\theta(1, \mathbf{x}_0) \approx \mathbf{x}_1 \sim p_1$.
In contrast to flow matching, which learns a velocity field
$v_\theta(\mathbf{x}_t, t)$ and requires numerical ODE integration at
inference, NFM learns the transport map itself.
This yields three key advantages: (1)~generation reduces to a single forward
pass $F_\theta(1, \mathbf{x}_0)$ with no ODE solver; (2)~training is
supervised directly on endpoint pairs $(\mathbf{x}_0, \mathbf{x}_1)$ rather
than instantaneous velocities; and (3)~any interpolation scheme — linear or
nonlinear — can be used to define intermediate supervision without affecting
the inference-time parameterization and efficiency.

\subsection{Naive Neural Flow Matching}
\label{subsec:naive_nf_matching}

A natural but naive approach to training neural flows is to regress
the flow map $F_\theta(t,\mathbf{x}_0)$ toward intermediate interpolation
points between \emph{independently} sampled source and target points,
i.e.\ $\pi := p_0 \times p_1$. Concretely, one samples
$\mathbf{x}_0 \sim p_0$, $\mathbf{x}_1 \sim p_1$ independently,
draws $t \sim \mathcal{U}(0,1)$, and defines the target state via
linear interpolation $\mathbf{x}_t := (1-t)\mathbf{x}_0 + t\mathbf{x}_1$
\cite{lipman2023flow,liu2023flow,tong2024improving}.
The neural flow is then trained by minimizing
\begin{equation}
\label{eqn:ind}
\mathcal{L}_{\text{naive}}(\theta)
= \mathbb{E}_{\substack{\mathbf{x}_0 \sim p_0,\,\mathbf{x}_1 \sim p_1 \\ t \sim \mathcal{U}(0,1)}}
\!\left[\bigl\| F_\theta(t,\mathbf{x}_0) - \mathbf{x}_t \bigr\|^2\right].
\end{equation}
Despite its simplicity, this objective admits a degenerate solution
whenever $\mathbf{x}_0$ and $\mathbf{x}_1$ are sampled independently,
as formalized below.

\begin{theorem}[Mean Collapse]
\label{thm:mean_collapse}
Let $F_\theta$ be an expressive neural flow trained under
$\mathcal{L}_{\mathrm{naive}}$ with independent coupling
$\pi = p_0 \times p_1$ and linear interpolation
$\mathbf{x}_t = (1-t)\mathbf{x}_0 + t\mathbf{x}_1$.
Then the unique minimizer satisfies
\begin{equation}
    F_{\theta^*}(t, \mathbf{x}_0)
    = (1-t)\mathbf{x}_0 + t\,\mathbb{E}_{p_1}[\mathbf{x}_1],
    \qquad \forall\, t \in [0,1],\; \mathbf{x}_0 \in \mathrm{supp}(p_0).
\end{equation}
In particular, $F_{\theta^*}(1, \mathbf{x}_0) = \mathbb{E}_{p_1}[\mathbf{x}_1]$
for all $\mathbf{x}_0$, collapsing every generated sample to the data mean.
\end{theorem}

% \begin{proof}
% For a squared loss, the pointwise minimizer at each $(t, \mathbf{x}_0)$
% is the conditional expectation:
% \begin{align}
% F_{\theta^*}(t, \mathbf{x}_0)
% &= \mathbb{E}\bigl[\mathbf{x}_t \mid t,\, \mathbf{x}_0\bigr] \notag\\
% &= \mathbb{E}\bigl[(1-t)\mathbf{x}_0 + t\mathbf{x}_1
%    \mid t,\, \mathbf{x}_0\bigr] \notag\\
% &= (1-t)\mathbf{x}_0 + t\,\mathbb{E}[\mathbf{x}_1 \mid \mathbf{x}_0] \notag\\
% &= (1-t)\mathbf{x}_0 + t\,\mathbb{E}_{p_1}[\mathbf{x}_1],
% \end{align}
% where the last step uses $\mathbf{x}_1 \perp \mathbf{x}_0$
% under $\pi = p_0 \times p_1$.
% \end{proof}

Theorem~\ref{thm:mean_collapse} reveals a fundamental failure mode
unique to neural flows: because $F_\theta$ conditions only on the
\emph{fixed} initial point $\mathbf{x}_0$, the squared-loss objective
drives all trajectories toward the global data mean as $t \to 1$,
rather than transporting mass toward the full target distribution.
This phenomenon arises for any interpolation trajectory (see \cref{eq:general_interp} whenever the coupling is independent.
We term this \emph{mean collapse}, and note that it has no analogue
in velocity-based flow matching: there, the model conditions on the
evolving state $\mathbf{x}_t$ at each ODE step and can self-correct
inconsistent pairings during integration.
Mean collapse therefore motivates replacing the independent coupling
with a structured pairing that assigns each $\mathbf{x}_0$ a
consistent target throughout training, which we develop in the
following section.

\subsection{Optimal Transport Neural Flow Matching}
\label{subsec:ot_coupling}

The mean collapse stems from
pairing inconsistency: the same $\mathbf{x}_0$ is matched to different targets
$\mathbf{x}_1$ across training steps, producing contradictory gradients that
cancel to the data mean.
Our proposal is to resolve this by replacing the independent coupling $\pi = p_0 \times p_1$
with the (deterministic) optimal transport plan:
\begin{equation}
     \pi^* = \argmin_{\pi \in \Pi(p_0,\, p_1)}
    \mathbb{E}_{(\mathbf{x}_0, \mathbf{x}_1) \sim \pi}
    \bigl[\|\mathbf{x}_0 - \mathbf{x}_1\|^2\bigr],
    \label{eq:ot_plan}   
\end{equation}

which assigns each $\mathbf{x}_0^{(i)}$ a partner
$\mathbf{x}_1^{(\sigma^*(i))}$ that remains consistent throughout training under the plan $\pi^*$.
The training objective is otherwise unchanged from ~\cref{eqn:ind}:
\begin{equation}
    \mathcal{L}_{\mathrm{OT}}(\theta)
    = \mathbb{E}_{(\mathbf{x}_0,\mathbf{x}_1)\sim\pi^*,\;t\sim\mathcal{U}(0,1)}
    \Bigl[\bigl\|F_\theta(t,\mathbf{x}_0) - \mathbf{x}_t\bigr\|^2\Bigr].
    \label{eq:ot_loss}
\end{equation} 

\begin{theorem}[OT Coupling is Necessary for Non-Degenerate Generation]
\label{thm:ot_necessity}
Let $F_\theta$ be an expressive neural flow trained under
$\mathcal{L}_{\mathrm{OT}}$ with coupling $\pi \in \Pi(p_0, p_1)$
and interpolation target
$\mathbf{x}_t = \alpha(t)\mathbf{x}_0 + \beta(t)\mathbf{x}_1$
for deterministic schedules $\alpha, \beta$ satisfying
$\alpha(0) = \beta(1) = 1$ and $\alpha(1) = \beta(0) = 0$.
The minimizer satisfies
\begin{equation}
    F_{\theta^*}(t, \mathbf{x}_0)
    = \alpha(t)\mathbf{x}_0 + \beta(t)\,\mathbb{E}_{\pi}[\mathbf{x}_1 \mid \mathbf{x}_0].
\end{equation}
The minimizer is non-degenerate, i.e.\
$F_{\theta^*}(1, \mathbf{x}_0) \not\equiv \mathbb{E}_{p_1}[\mathbf{x}_1]$,
if and only if $\pi \neq p_0 \times p_1$.
\end{theorem}

\paragraph{Coupling strategies.} Theorem~\ref{thm:ot_necessity} establishes 
that any non-independent coupling suffices for non-degenerate generation, but 
leaves open the question of which coupling is practical at scale. Exact global OT on $N$ samples requires solving an $N \times N$ linear program
($\mathcal{O}(N^3)$), which becomes intractable for large  or high dimensional datasets such as images. We consider three practical alternatives.
\emph{Minibatch OT}~\cite{tong2024improving} precomputes an approximate global
assignment by solving sequential non-overlapping $B \times B$ subproblems over
the dataset; since each subproblem yields an exact permutation matrix, the
assignment $\hat{\sigma}(i) = j^*$ is read directly from each chunk's solution
and fixed for all training steps. Another strategy is 
\emph{LOOM} where we ~\cite{davtyan2025faster} maintain a persistent coupling $\tau$
initialized to the identity and refines it online via minibatch OT at each
training step, gradually converging to the global plan without any
precomputation. The third rather crude strategy is 
\emph{Per-batch OT}, which solves a fresh $B \times B$ problem at each step, providing
local optimality within each minibatch but no global consistency.% across steps. %We evaluate all three strategies in our experiments.

\subsection{Neural Flow Models}
\label{subsec:nf_models}

Different neural architectures yield different instantiations of the
flow map $F_\theta$. Our primary model is \emph{ResNetFlow}~\cite{bilovs2021neural},
which parameterizes the flow map as
\begin{equation}
    F_\theta(t, \mathbf{x}_0) = \mathbf{x}_0 + \phi(t)\,g_\theta(t, \mathbf{x}_0),
    \label{eq:resnetflow}
\end{equation}
where $g_\theta$ is a spectrally normalized residual
network~\cite{gouk2021regularisation} operating on
$[\mathbf{x}_0 \| \phi(t)]$, enforcing $\mathrm{Lip}(g_\theta) < 1$
and guaranteeing that $F_\theta(t, \cdot)$ is a bijection for all $t$.
The identity condition $F_\theta(0, \mathbf{x}_0) = \mathbf{x}_0$ is
satisfied by construction via $\phi(0) = 0$.
While other neural flow architectures exist, such as \emph{CouplingFlow}
and \emph{GRUFlow}~\cite{bilovs2021neural}, these are designed primarily
for irregularly-sampled time series modeling and are not directly
applicable to the generative flow map setting considered here.
We therefore adopt ResNetFlow as our backbone throughout all experiments.

\begin{proposition}[Time-Independent Displacement under Linear Interpolation]
\label{prop:linear_optimal}
Let $F_\theta(t, \mathbf{x}_0) = \mathbf{x}_0 + \phi(t)\,g_\theta(t, \mathbf{x}_0)$
be a ResNetFlow with $\phi(t) = t$, trained under $\mathcal{L}_{\mathrm{OT}}$
with a deterministic OT coupling and linear interpolation
$\mathbf{x}_t = (1-t)\mathbf{x}_0 + t\mathbf{x}_1$.
Then the optimal displacement network satisfies
\begin{equation}
\label{eq:g_optimal}
    g^*(t, \mathbf{x}_0) = \mathbf{x}_1 - \mathbf{x}_0,
\end{equation}
which is time-independent. For any $\phi(t) \not\propto t$, the optimal
$g^*$ depends explicitly on $t$, strictly increasing learning complexity.
\end{proposition}

% \begin{proof}
% Under a deterministic OT coupling, $\mathbf{x}_1 = T(\mathbf{x}_0)$\;
% $\pi$-a.s., so $\mathbb{E}_\pi[\mathbf{x}_1 \mid \mathbf{x}_0] = T(\mathbf{x}_0)$.
% The minimizer of $\mathcal{L}_{\mathrm{OT}}$ must satisfy
% \begin{equation}
%     \mathbf{x}_0 + t\,g^*(t, \mathbf{x}_0)
%     = (1-t)\mathbf{x}_0 + t\,T(\mathbf{x}_0).
% \end{equation}
% Solving for $g^*$:
% \begin{equation}
% \label{eq:g_optimal}
%     g^*(t, \mathbf{x}_0)
%     = \frac{(1-t)\mathbf{x}_0 + t\,T(\mathbf{x}_0) - \mathbf{x}_0}{t}
%     = T(\mathbf{x}_0) - \mathbf{x}_0
%     = \mathbf{x}_1 - \mathbf{x}_0,
% \end{equation}
% which is independent of $t$. For any $\phi(t) \not\propto t$, the
% $t$-dependence does not cancel and $g^*$ must additionally model
% temporal variation, increasing learning complexity.
% \end{proof}

\begin{remark}
Proposition~\ref{prop:linear_optimal} has two important implications.
First, $g_\theta$ need only learn a static spatial displacement field
$\mathbf{x}_0 \mapsto \mathbf{x}_1 - \mathbf{x}_0$, independent of $t$ —
the simplest possible regression target for a neural flow.
Second, $g_\theta$ does not observe $\mathbf{x}_1$ at inference and must
recover the displacement from $\mathbf{x}_0$ alone, which is well-posed
only when pairings are globally consistent.
Without OT, $g^*$ degenerates to $\mathbb{E}_{p_1}[\mathbf{x}_1] - \mathbf{x}_0$,
recovering exactly the mean collapse of
Theorem~\ref{thm:mean_collapse}.
These two properties jointly justify our design choices of linear
interpolation and OT coupling, and explain why linear interpolation
consistently outperforms all other trajectory schemes in our
ablation (Section~\ref{subsubsec:traj_ablation}).
\end{remark}

\paragraph{Network architecture.}
For 2D data, $g_\theta$ is instantiated as a spectrally
normalized~\cite{miyato2018spectral} residual network:
\begin{equation}
    \mathbf{h}^{(l+1)} = \mathbf{h}^{(l)}
    + W_2^{(l)}\,\sigma\!\left(W_1^{(l)}\mathbf{h}^{(l)}\right),
    \qquad \mathbf{h}^{(0)} = [\mathbf{x}_0 \,\|\, \phi(t)].
    \label{eq:resnet_layers}
\end{equation}
%Spectral normalization enforces $\mathrm{Lip}(g_\theta) < 1$,
%guaranteeing $F_\theta(t, \cdot)$ is a bijection for all
%$t$~\cite{bilovs2021neural}.
%For image data $\mathbf{x}_0 \in \mathbb{R}^{C \times H \times W}$, $g_\theta$ is instantiated as a UNet~\cite{ronneberger2015unet} that
%produces a pixel-wise displacement map of the same spatial dimensions
%as $\mathbf{x}_0$, with time embedding $\tau(t) = \mathrm{MLP}(\mathrm{sinusoid}(t))$
%injected at every resolution via adaptive group
%normalization~\cite{dhariwal2021diffusion}.
\paragraph{UNet displacement for image data.}
For image data $\mathbf{x}_0 \in \mathbb{R}^{C \times H \times W}$,
we instantiate $g_\theta$ as a UNet~\cite{ronneberger2015unet} that
produces a pixel-wise displacement map of the same spatial dimensions
as $\mathbf{x}_0$, which we call \emph{ResNetFlowUNet}.
The time embedding $\tau(t) = \mathrm{MLP}(\mathrm{sinusoid}(t))$ is
injected at every resolution via adaptive group
normalization~\cite{dhariwal2021diffusion}, allowing $g_\theta$ to
modulate displacement magnitude smoothly with $t$.
Unlike the flat residual network, the UNet's convolutional weight
sharing generalizes across spatial locations: $g_\theta$ learns that
patches with a given appearance should receive a given displacement,
a property that transfers naturally to unseen noise inputs at inference.
While the UNet relaxes the spectral normalization constraint and
therefore does not formally guarantee invertibility of $F_\theta$,
this is not required for generation. % --- only the identity condition$F_\theta(0, \mathbf{x}_0) = \mathbf{x}_0$, enforced by $\phi(0) = 0$, is necessary.

\subsection{Interpolation Trajectories}
\label{subsec:trajectories}

A key design choice in NFM is the interpolation scheme used to define the
target trajectory $\mathbf{x}_t$ between source $\mathbf{x}_0$ and target
$\mathbf{x}_1$.
Unlike flow matching, which learns a velocity field $v_\theta(\mathbf{x}_t,t)$
that adapts to the current state at each ODE step, neural flows learn a map
$F_\theta(t, \mathbf{x}_0)$ conditioned only on the \emph{initial} state
$\mathbf{x}_0$.
Crucially, this means NFM is not restricted to straight-line trajectories:
since $F_\theta$ directly models a curve in $\mathbb{R}^d$ parameterized by
$t$, it can represent any smooth path between $\mathbf{x}_0$ and $\mathbf{x}_1$
without the trajectory curvature penalty that afflicts ODE-based methods at
inference.

Given a (OT)-coupled pair $(\mathbf{x}_0, \mathbf{x}_1) \sim \pi$, we consider a
general family of interpolation schemes:
\begin{equation}
    \mathbf{x}_t = \alpha(t)\,\mathbf{x}_0 + \beta(t)\,\mathbf{x}_1
    + \sigma(t)\,\boldsymbol{\varepsilon},
    \quad \boldsymbol{\varepsilon} \sim \mathcal{N}(\mathbf{0}, \mathbf{I}),
    \label{eq:general_interp}
\end{equation}
where $\alpha(t), \beta(t)$ are deterministic schedules satisfying
$\alpha(0) = \beta(1) = 1$ and $\alpha(1) = \beta(0) = 0$, and $\sigma(t)
\geq 0$ is an optional noise schedule.
Table~\ref{tab:trajectories} summarizes the specific instantiations considered
in this work, along with their induced velocities and key properties.

\paragraph{Implications for neural flow matching.}
The choice of trajectory directly determines the difficulty of learning
$g_\theta$ in ResNetFlow (\cref{eq:resnetflow}) or its variant ResNetFlowUNet for images.
For the \textbf{linear} scheme, the velocity $\dot{\mathbf{x}}_t =
\mathbf{x}_1 - \mathbf{x}_0$ is constant in time, so $g_\theta$ need only
learn a time-independent spatial displacement field as shown by the optimal
solution in~\cref{eq:g_optimal}.
This makes linear interpolation the most natural and efficient choice for
ResNetFlow.
Schemes with time-varying velocity (\textbf{cosine}, \textbf{polynomial}) require $g_\theta$ to additionally model temporal variation,
increasing learning difficulty without benefiting one-step generation.
\textbf{Stochastic} trajectories add noise to the target, which regularizes
training against mean collapse in the absence of OT pairing, but introduce
endpoint variance that degrades one-step generation quality.
%The \textbf{Schrödinger bridge} scheme combines both time-varying and stochastic velocity, making it poorly suited for direct flow map regression.
In summary, the geometric flexibility of neural flows, their ability to model
arbitrary curves, is most effectively exploited with simple, straight
trajectories that reduce the temporal learning burden on $g_\theta$.

\begin{table*}[hbtp]
\centering
\caption{
Interpolation and probability paths used in neural flow matching.
Deterministic paths connect two fixed endpoints
$\mathbf{x}_0 \rightarrow \mathbf{x}_1$,
while stochastic paths connect probability distributions.
%Const.\ vel.\ indicates time-independent velocity.
%Exact endpoint indicates $\mathbf{x}_{t=1}=\mathbf{x}_1$ deterministically.
}
\label{tab:trajectories}
\resizebox{\textwidth}{!}{%
\begin{tabular}{llllcc}
\toprule
\textbf{Trajectory}
& \textbf{Path $\mathbf{x}_t$}
& \textbf{Velocity $\dot{\mathbf{x}}_t$}
& \textbf{Const.\ vel.}
& \textbf{Exact endpoint}
& \textbf{Reference} \\
\midrule

% \multicolumn{6}{l}{\textbf{Deterministic interpolants}} \\
% \midrule

Linear
& $(1-t)\mathbf{x}_0 + t\mathbf{x}_1$
& $\mathbf{x}_1-\mathbf{x}_0$
& \checkmark
& \checkmark
& \cite{lipman2023flow, liu2023flow} \\[4pt]

Cosine
& $\cos\!\left(\tfrac{\pi t}{2}\right)\mathbf{x}_0
  + \sin\!\left(\tfrac{\pi t}{2}\right)\mathbf{x}_1$
& $\tfrac{\pi}{2}
   \!\left[-\sin\!\left(\tfrac{\pi t}{2}\right)\mathbf{x}_0
   + \cos\!\left(\tfrac{\pi t}{2}\right)\mathbf{x}_1\right]$
& $\times$
& \checkmark
& \cite{chen2023importance} \\[4pt]

Polynomial ($\alpha$)
& $(1-t^\alpha)\mathbf{x}_0 + t^\alpha\mathbf{x}_1$
& $\alpha t^{\alpha-1}(\mathbf{x}_1-\mathbf{x}_0)$
& $\times$
& \checkmark
& --- \\[8pt]

% \midrule
% \multicolumn{6}{l}{\textbf{Stochastic probability paths}} \\
% \midrule

% VP diffusion
% & $\alpha(t)\mathbf{x}_0 + \sigma(t)\boldsymbol{\varepsilon}$,
%   $\ \sigma^2(t)=1-\alpha^2(t)$
% & $\dot{\alpha}(t)\mathbf{x}_0
%    + \dot{\sigma}(t)\boldsymbol{\varepsilon}$
% & $\times$
% & $\times$
% & \cite{ho2020denoising, song2021score} \\[6pt]

Stochastic interpolant
& $(1-t)\mathbf{x}_0 + t\mathbf{x}_1
  + \sigma\sqrt{t(1-t)}\,\boldsymbol{\varepsilon}$
& $\mathbf{x}_1-\mathbf{x}_0
  + \tfrac{\sigma(1-2t)}{2\sqrt{t(1-t)}}\boldsymbol{\varepsilon}$
& $\times$
& $\times$
& \cite{albergo2023stochastic} \\[6pt]

% Schr\"odinger bridge (Gaussian case)
% & mean path as linear interpolation,
%   covariance solves SB equations
% & depends on score functions
% & $\times$
% & $\times$
% & \cite{debortoli2023diffusion, tong2024improving} \\

\bottomrule
\end{tabular}%
}
\end{table*}

\section{Experiments}
\label{sec:experiments}

\subsection{Synthetic Experiments}
\label{sec:synthetic}

We evaluate OT-NFM on four 2-D transport benchmarks designed to probe how
coupling quality affects trajectory consistency, multimodal coverage, and
the stability of the learned flow map.
%All models use \textsc{ResNetFlow} with linear interpolation ($\phi(t)=t$),
%trained for 5{,}000 steps with batch size 128 on $N=1{,}000$ samples.
We compare four coupling strategies: \emph{per-batch OT}, \emph{minibatch OT},
\emph{LOOM}~\cite{davtyan2025faster}, and \emph{global OT}.

\paragraph{Benchmarks.}
\textbf{Gauss\,$\to$\,Checkerboard.}
The source is $\mathcal{N}(\mathbf{0},\mathbf{I})$ and the target is a
checkerboard of 8 alternating tiles on $[-4,4]^{2}$.
The disconnected target support makes pairing errors immediately visible:
mass transported to the wrong tile cannot be corrected at inference.
\textbf{Gauss\,$\to$\,Spiral.}
The target is a two-dimensional Archimedean spiral embedded in a noisy band.
This benchmark tests whether the flow map can learn coherent long-range,
curved routing from a unimodal source.
\textbf{Gauss\,$\to$\,Crescent.}
The target is a crescent-shaped region requiring the model to learn a
strongly asymmetric, non-convex displacement field from a symmetric source.
\textbf{8-GMM\,$\to$\,2-Moons.}
The source is an 8-component Gaussian mixture arranged on a circle and the
target is the two-moons distribution.
This task stresses long-range, multi-to-multi routing: each of the 8 source
modes must be stably assigned to one of the two crescents throughout training.
\begin{figure}[h!]
  \centering
  \includegraphics[width=0.9\linewidth]{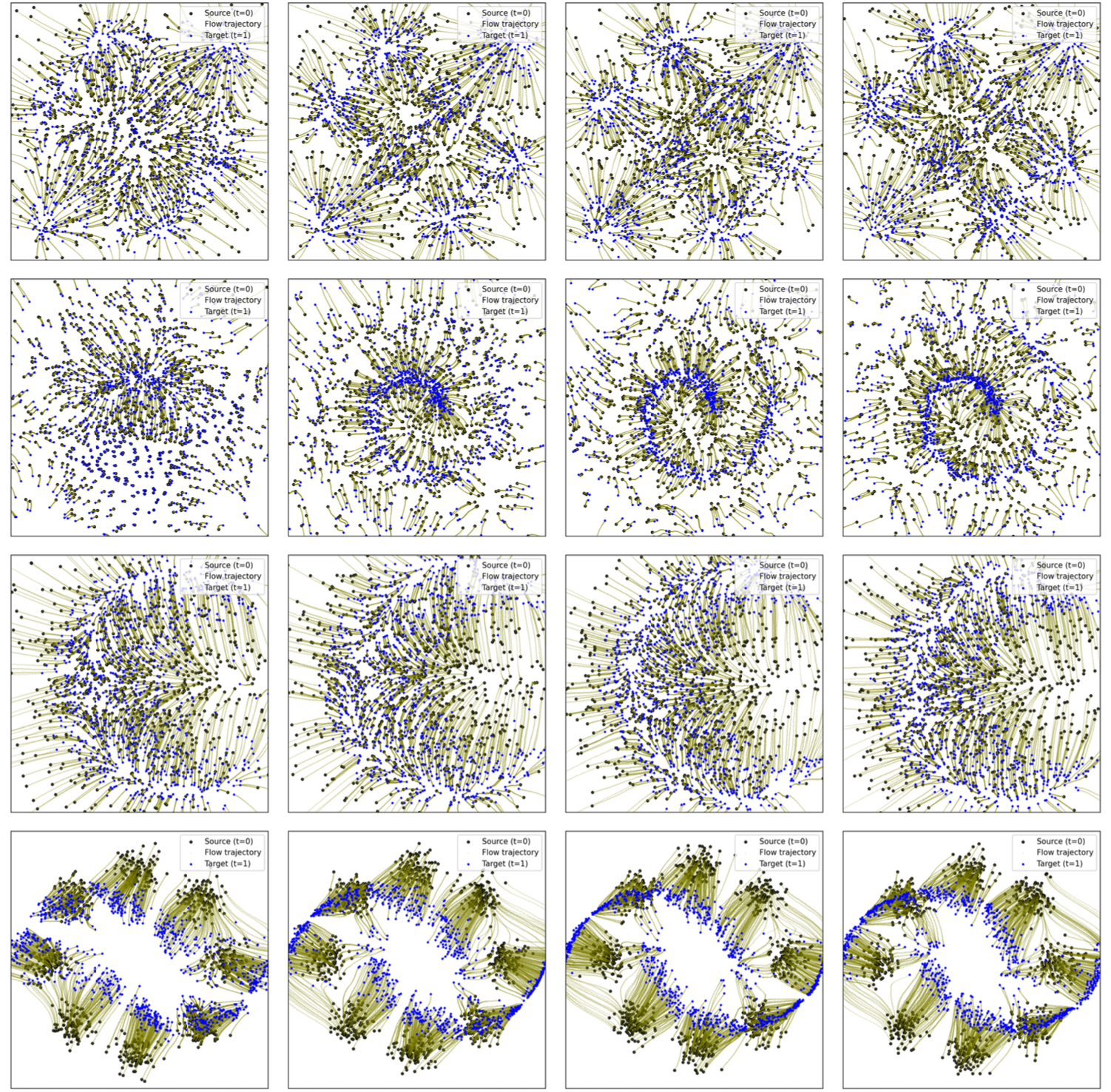}
  \caption{%
    \textbf{Synthetic transport results.}
    Flow trajectories under four coupling strategies (columns) on four 2-D
    benchmarks (rows): Gauss\,$\to$\,Checkerboard, Gauss\,$\to$\,Spiral,
    Gauss\,$\to$\,Crescent, and 8-GMM\,$\to$\,2-Moons.
    Black dots: source samples ($t=0$); blue dots: generated samples ($t=1$);
    olive arrows: learned flow trajectories.
    \emph{Per-batch OT} produces tangled, inconsistent trajectories across all
    tasks.
    \emph{Minibatch OT}, \emph{LOOM}, and \emph{Global OT} all recover clean
    transport with coherent, largely non-crossing routing, confirming that
    globally consistent coupling is the key determinant of flow-map quality.%
  }
  \label{fig:synthetic}
\end{figure}

\paragraph{Results.}
Figure~\ref{fig:synthetic} shows the learned flow trajectories for all four
tasks and all four coupling strategies.
The results reveal a consistent hierarchy in trajectory quality that is
determined almost entirely by the global consistency of the source–target pairing.

\emph{Global OT} produces the sharpest results across all tasks.
On Gauss\,$\to$\,Checkerboard, radial, non-crossing trajectories route each
source region to its geometrically nearest tile, yielding crisp tile boundaries.
On Gauss\,$\to$\,Spiral and Gauss\,$\to$\,Crescent, trajectories fan out
coherently from the source, smoothly filling the curved and asymmetric target
supports.
On 8-GMM\,$\to$\,2-Moons, each source mode is committed to a single crescent
with no inter-mode crossings.

\emph{Minibatch OT} achieves near-identical coverage and trajectory structure
on all four tasks.
Because each chunk's local assignment is an exact permutation matrix and is
fixed for all training steps, the flow map receives gradient signals that are
globally consistent up to the granularity of the chunk size.
The only visible deviation from global OT is a mild increase in crossing
trajectories on the checkerboard, reflecting the approximate nature of
non-overlapping chunk assignments.

\emph{LOOM} matches or closely approaches minibatch OT on every benchmark.
Its online refinement of the coupling converges to a globally consistent
assignment over training, yielding clean spiral and crescent coverage
and well-separated crescent routing on 8-GMM\,$\to$\,2-Moons.
LOOM offers a practical middle ground: it requires no precomputation, yet its
iterative refinement eliminates the per-batch inconsistency that degrades
trajectory quality.

\emph{Per-batch OT} consistently produces the worst trajectories.
Without a persistent assignment, the same source point is coupled to different
targets across batches, generating contradictory gradients that pull the
displacement network in conflicting directions.
The effect is most pronounced on 8-GMM\,$\to$\,2-Moons, where the two
crescents should be sharply separated but trajectories from the same source
mode fan out toward both, producing tangled, inconsistent paths.
On simpler tasks (Gauss\,$\to$\,Spiral, Gauss\,$\to$\,Crescent) per-batch OT
partially recovers the target support, but with visibly higher trajectory
variance than the globally consistent strategies.

% \paragraph{Takeaway.}
% Across all four tasks, globally consistent coupling is the single most
% important factor for trajectory quality in neural flow matching.
% This stands in contrast to velocity-based flow matching methods such as OT-CFM,
% which self-correct at each ODE step and are therefore robust to pairing quality.
% Because neural flows condition only on the fixed initial noise $\mathbf{x}_{0}$
% and have no mechanism to adapt mid-trajectory, the displacement network $g_\theta$
% must receive consistent gradient signals throughout training.
% Either precomputed minibatch OT or online refinement via LOOM provides this
% consistency at a fraction of the cost of global OT, making both practical choices
% for large-scale training.

% \begin{table}[t]
% \centering
% \caption{$W_2^2$ on Gaussian $\to$ 2-Moons ($\mu \pm \sigma$, 5 seeds).
% OT-NFM achieves the best W$_2^2$ with a single forward pass and no ODE solver.}
% \label{tab:w2_moons}
% \begin{tabular}{lcc}
% \toprule
% Method & NFE $\downarrow$ & $W_2^2$ $\downarrow$ \\
% \midrule
% iCFM~\cite{lipman2023flow}        & 100 & 0.0793 \scriptsize{$\pm$ 0.0217} \\
% MeanFlow~\cite{geng2025mean}      &   1 & 0.0557 \scriptsize{$\pm$ 0.0119} \\
% OT-CFM~\cite{tong2024improving}   & 100 & 0.0324 \scriptsize{$\pm$ 0.0082} \\
% \midrule
% OT-NFM (ours)                     &   1 & \textbf{0.0230} \scriptsize{$\pm$ \textbf{0.0046}} \\
% \bottomrule
% \end{tabular}
% \end{table}

\paragraph{Synthetic $W_2^2$ evaluation.}
We quantitatively evaluate distribution matching quality using the squared 2-Wasserstein distance ($W_2^2$) following \cite{tong2024improving} on two 2D benchmarks: Gaussian $\to$ 2-Moons and 8-GMM $\to$ 2-Moons.
We compare OT-NFM against iCFM~\cite{lipman2023flow}, OT-CFM~\cite{tong2024improving}, and MeanFlow~\cite{geng2025mean}, reporting $\mu \pm \sigma$ over 5 random seeds.
Table~\ref{tab:w2} summarizes the results.
On Gaussian $\to$ Moons, OT-NFM achieves the lowest $W_2^2$ of all methods, including OT-CFM which requires 100$\times$ more function evaluations.
On the harder 8-GMM $\to$ Moons benchmark, OT-NFM remains statistically competitive with OT-CFM while using a single forward pass.
In both settings, OT-NFM substantially outperforms MeanFlow, the only other 1-NFE baseline, demonstrating that OT coupling --- not just one-step generation --- is the key ingredient for distribution matching quality.
iCFM degrades severely on the structured 8-GMM source ($W_2^2 = 0.1677$), confirming that independent coupling is insufficient when both the source and target distribution are multimodal.

\begin{table}[t]
\centering
\caption{$W_2^2$ ($\mu \pm \sigma$, 5 seeds) on two 2D benchmarks.
\textbf{Bold}: best overall. \textit{Italic}: best among 100-NFE methods (OT-CFM) or best among 1-NFE methods (OT-NFM).
OT-NFM matches or surpasses OT-CFM with a single forward pass and no ODE solver.}
\label{tab:w2}
\begin{tabular}{lc cc}
\toprule
 & & \multicolumn{2}{c}{$W_2^2$ $\downarrow$} \\
\cmidrule(lr){3-4}
Method & NFE $\downarrow$ & Gaussian $\to$ Moons & 8-GMM $\to$ Moons \\
\midrule
iCFM~\cite{lipman2023flow}       & 100 & 0.0793 \scriptsize{$\pm$ 0.0217} & 0.1677 \scriptsize{$\pm$ 0.0524} \\
MeanFlow~\cite{geng2025mean}     &   1 & 0.0557 \scriptsize{$\pm$ 0.0119} & 0.0574 \scriptsize{$\pm$ 0.0130} \\
OT-CFM~\cite{tong2024improving}  & 100 & \textit{0.0324} \scriptsize{$\pm$ 0.0082} & \textit{0.0323} \scriptsize{$\pm$ 0.0062} \\
\midrule
OT-NFM (ours)                    &   1 & \textbf{0.0230} \scriptsize{$\pm$ 0.0046} & \textit{0.0337} \scriptsize{$\pm$ 0.0166} \\
\bottomrule
\end{tabular}
\end{table}

\subsection{Image Generation}
\label{subsec:image}
% ============================================================

% \subsubsection{MNIST}
% \label{subsubsec:mnist}

% We train NeuralFlowUNet on MNIST ($28{\times}28$, grayscale) using precomputed
% minibatch OT on for $2$ steps.
% Since FID is unreliable for $28{\times}28$ grayscale images, we report visual
% sample quality and coupling ablation.

% \paragraph{Coupling ablation.}
% Figure~\ref{fig:mnist_ablation} shows samples from NF and all baselines under
% each coupling strategy.
% OT-CFM, MeanFlow, and Flow Map Matching all produce recognizable digits
% regardless of coupling, confirming robustness via self-correction.
% NF without coupling produces blurred averages consistent with mean collapse
% (Section~\ref{subsec:naive_nfm}); precomputed minibatch OT restores sharp,
% diverse digit generation.

% % [FIGURE: rows = methods (NF×4 couplings, OT-CFM, MeanFlow, FMM),
% %  cols = samples. Caption highlights collapse row vs sharp rows.]
% \begin{figure}[h]
% \centering
% % \includegraphics[width=\linewidth]{figures/mnist_ablation.pdf}
% \caption{
%   \textbf{MNIST samples.}
%   Rows: NF (no coupling / per-batch / minibatch OT / global OT),
%   OT-CFM, MeanFlow, Flow Map Matching.
%   NF without coupling collapses; precomputed OT restores sharp digits.
%   All baselines are robust to coupling choice.
% }
% \label{fig:mnist_ablation}
% \end{figure}
\subsubsection{MNIST}
\label{subsubsec:mnist}

We train our neural flow model on MNIST ($28{\times}28$, grayscale) with
precomputed minibatch OT.
Since FID is unreliable for $28{\times}28$ grayscale images, we assess
quality through visual inspection and a coupling ablation.

\begin{figure}[t]
  \centering
  \includegraphics[width=0.195\linewidth]{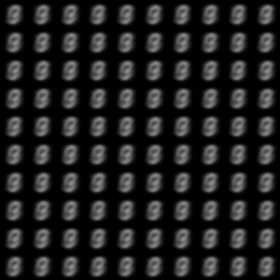}\hfill
  \includegraphics[width=0.195\linewidth]{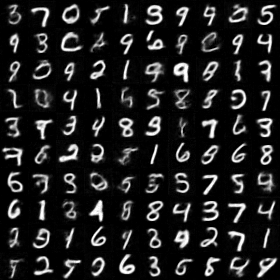}\hfill
  \includegraphics[width=0.195\linewidth]{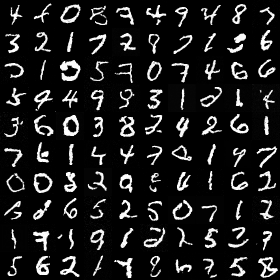}\hfill
  \includegraphics[width=0.195\linewidth]{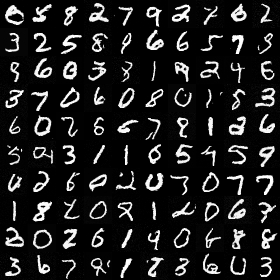}\hfill
  \includegraphics[width=0.195\linewidth]{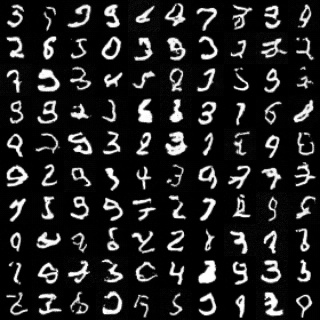}
  \\[2pt]
  \begin{tabular}{@{}p{0.195\linewidth}p{0.195\linewidth}p{0.195\linewidth}p{0.195\linewidth}p{0.195\linewidth}@{}}
    \centering\small NFM (no OT) &
    \centering\small NFM (minibatch OT) &
    \centering\small CFM (no OT) &
    \centering\small CFM (minibatch OT) &
    \centering\small MeanFlow
  \end{tabular}
  \caption{%
    \textbf{MNIST coupling ablation.}
    Each panel shows 100 generated samples ($10{\times}10$ grid).
    \emph{NFM (no OT):} complete mean collapse --- every output is a blurred
    average with no digit identity.
    \emph{NFM (minibatch OT):} precomputed pairing restores sharp, diverse
    digits in a single forward pass.
    \emph{CFM (no OT)} provides an unguided flow matching baseline with
    independent coupling.
    \emph{CFM (minibatch OT)} (100 NFE) and \emph{MeanFlow} (1 NFE) produce
    recognizable digits regardless of coupling, confirming robustness via
    iterative self-correction.
  }
  \label{fig:mnist_ablation}
\end{figure}

\paragraph{Coupling ablation.}
Figure~\ref{fig:mnist_ablation} compares samples across methods and coupling
strategies.
The no-coupling baseline (NF, no OT) collapses entirely to a blurred, nearly
identical blob for every noise input which is a direct visual confirmation of the mean
collapse described in \cref{subsec:naive_nf_matching}.
Replacing the independent coupling with precomputed minibatch OT immediately
restores sharp, diverse digit generation, demonstrating that consistent
source--target pairing is both necessary and sufficient to avoid collapse.

% Among the baselines, OT-CFM produces the sharpest strokes overall, benefiting
% from $100$ NFE of iterative correction.
% Our minibatch OT model generates comparably legible digits in a \emph{single
% forward pass}, with no ODE solver.
% MeanFlow digits are recognizable but exhibit thicker, less precise strokes,
% suggesting that its average-velocity parameterization introduces mild blurring
% at this scale and training budget.
% Importantly, all velocity-based baselines (OT-CFM, MeanFlow) remain robust to
% coupling quality, since their iterative inference self-corrects pairing errors
% at each step, a robustness that neural flows cannot rely on. It is also worth noting that the training of MeanFlow for images rely on the Dissufuion transform models \cite{peebles2023scalable}. Hence model training time more doubles than the rest models. 
Among the baselines, OT-CFM produces the sharpest strokes overall, benefiting
from 100 NFE of iterative correction.
Our minibatch OT model generates comparably legible digits in a \emph{single
forward pass} with no ODE solver.
MeanFlow digits are recognizable but exhibit thicker, less precise strokes;
we attribute this to the fact that the original MeanFlow implementation for
image generation relies on Diffusion Transformer (DiT)
backbones~\cite{peebles2023scalable}, whereas our comparison uses the same
lightweight UNet architecture across all methods for a controlled evaluation.
Importantly, all velocity-based baselines (OT-CFM, MeanFlow) remain robust to
coupling quality, since iterative inference self-corrects pairing errors at
each step --- a robustness that neural flows cannot rely on.

% \begin{figure}[t]
%   \centering
%   \includegraphics[width=\linewidth]{figures/mnist_ablation}
%   \caption{%
%     \textbf{MNIST coupling ablation.}
%     Each panel shows 100 generated samples ($10{\times}10$ grid).
%     \emph{NF (no OT):} complete mean collapse — every output is a blurred
%     average with no digit identity.
%     \emph{NF (minibatch OT):} precomputed pairing restores sharp, diverse
%     digits in a single forward pass.
%     \emph{OT-CFM} (100 NFE) and \emph{MeanFlow} (1 NFE) produce recognizable
%     digits regardless of coupling, confirming robustness via iterative
%     self-correction.
%     \emph{iCFM} serves as an unguided flow matching baseline.
%   }
%   \label{fig:mnist_ablation}
% \end{figure}

% \paragraph{Trajectory visualization.}
% Figure~\ref{fig:mnist_traj} shows $F_\theta(t, \mathbf{x}_0)$ for fixed
% $\mathbf{x}_0$ across $t \in \{0, 0.1, \ldots, 1.0\}$.
% At $t{=}0$ the identity condition gives pure noise; digit structure emerges
% progressively with class committed by $t{\approx}0.5$.

\begin{figure}[t]
  \centering
  \includegraphics[width=\linewidth]{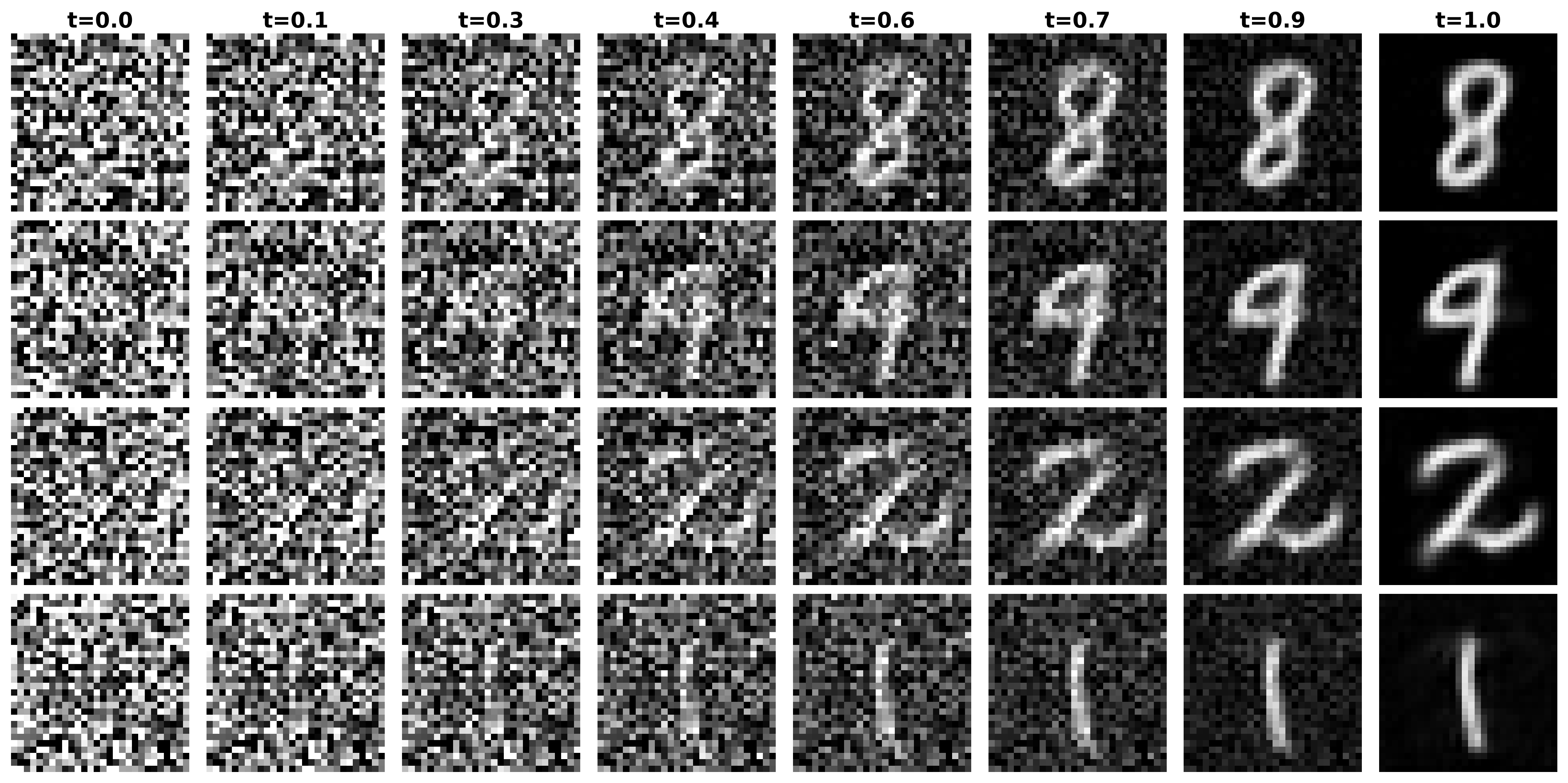}
  \caption{%
    \textbf{Neural flow trajectories on MNIST.}
    Each row shows $F_\theta(t, \mathbf{x}_0)$ for a fixed noise vector
    $\mathbf{x}_0 \sim \mathcal{N}(\mathbf{0}, \mathbf{I})$ evaluated at
    $t \in \{0.0, 0.1, \ldots, 1.0\}$.
    At $t{=}0$ the identity condition gives pure noise; coarse digit structure
    emerges by $t{\approx}0.3$--$0.4$ and class identity is committed by
    $t{\approx}0.5$; background noise is suppressed and strokes sharpen
    monotonically toward the clean output at $t{=}1.0$.
    No ODE solver is used; generation is a single forward pass of
    $F_\theta(1, \mathbf{x}_0)$.
  }
  \label{fig:mnist_traj}
\end{figure}

% % [FIGURE: 8 rows (samples) x 10 cols (t=0..1)]
% \begin{figure}[h]
% \centering
% % \includegraphics[width=\linewidth]{figures/mnist_traj.pdf}
% \caption{
%   \textbf{Neural flow trajectory on MNIST.}
%   $F_\theta(t, \mathbf{x}_0)$ for 8 fixed noise vectors across
%   $t \in \{0, 0.1, \ldots, 1.0\}$.
%   Identity at $t{=}0$; digit identity committed by $t{\approx}0.5$.
% }
% \label{fig:mnist_traj}
% \end{figure}
\paragraph{Trajectory visualization.}
Figure~\ref{fig:mnist_traj} shows $F_\theta(t, \mathbf{x}_0)$ for four fixed
noise vectors across $t \in \{0.0, 0.1, \ldots, 1.0\}$, evaluated in a single
forward pass with no ODE solver.
At $t{=}0$ the identity condition is satisfied exactly, giving pure noise.
Digit structure begins to emerge around $t{\approx}0.3$--$0.4$, with class
identity committed by $t{\approx}0.5$; the remaining steps progressively
sharpen stroke contrast and suppress background noise until a clean digit
is recovered at $t{=}1$.

% ============================================================
\subsubsection{CIFAR-10}
\label{subsubsec:cifar}

\begin{wrapfigure}{r}{0.4\linewidth}
  \centering
  \vspace{-10pt}
  \includegraphics[width=\linewidth]{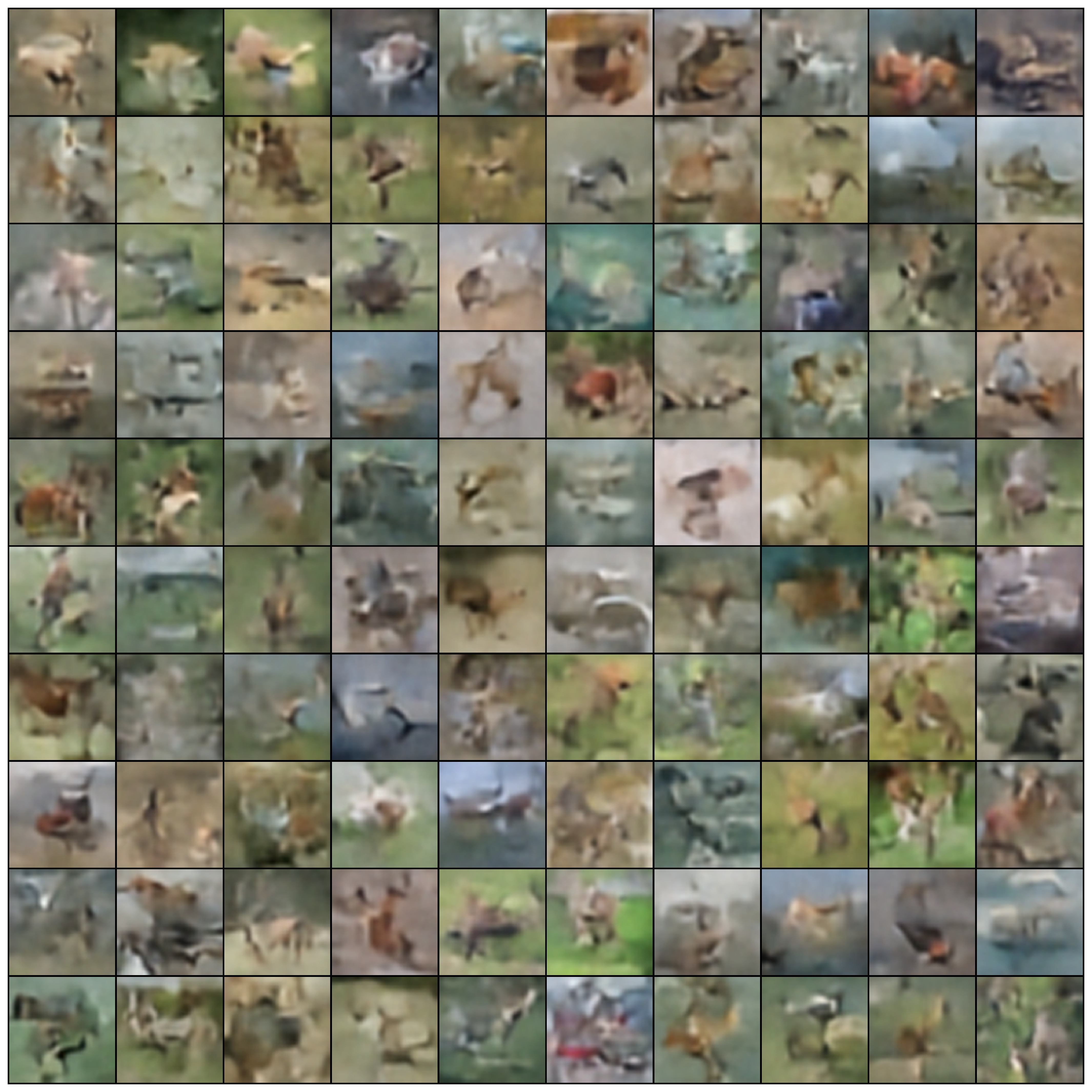}
  \caption{%
    \textbf{CIFAR-10 samples (1-NFE).}
  }
  \label{fig:cifar_samples}
\end{wrapfigure}

We train our neural flow model on CIFAR-10 ($32{\times}32$, RGB)
with precomputed minibatch OT ($N{=}50{,}000$, $B{=}256$, 5 sweep
epochs) for $100{,}000$ steps using the Adam optimizer with learning
rate $2{\times}10^{-4}$ and cosine annealing decay on an NVIDIA A100 GPU.
Figure~\ref{fig:cifar_samples} shows 100 generated samples from a
single forward pass $F_\theta(1, \mathbf{x}_0)$ with no ODE solver.
The model produces diverse samples with varied textures, poses, and
color palettes without mean collapse, confirming that precomputed
minibatch OT coupling scales from synthetic 2D distributions to
natural image data.
We note that stronger quantitative results would require EDM-style
preconditioning~\cite{karras2022elucidating} and larger backbone
architectures~\cite{peebles2023scalable}, which we leave for future work.

\subsection{Ablation Experiment: Trajectory Interpolation}
\label{subsubsec:traj_ablation}
We ablate four interpolation schemes across two benchmarks with global OT
coupling fixed, isolating the effect of the trajectory on learned flow map
quality. The linear trajectory (already shown in Figure~\ref{fig:synthetic})
serves as our reference in both cases.

\begin{figure}[h]
  \centering
  % Top row: 8-GMM → 2-moon
  % Top row: 8-GMM → 2-moons
  \includegraphics[width=0.3\linewidth]{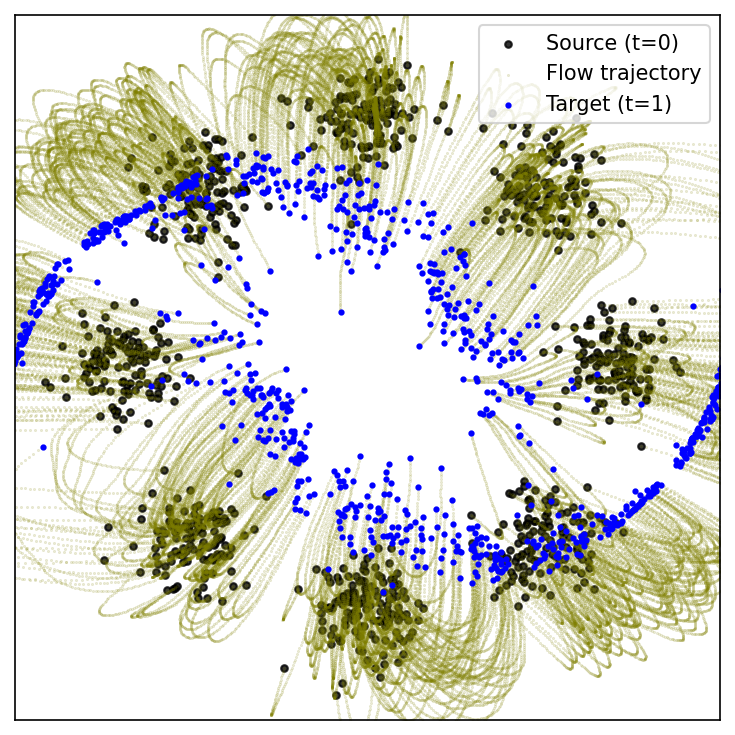}\hspace{2pt}
  \includegraphics[width=0.3\linewidth]{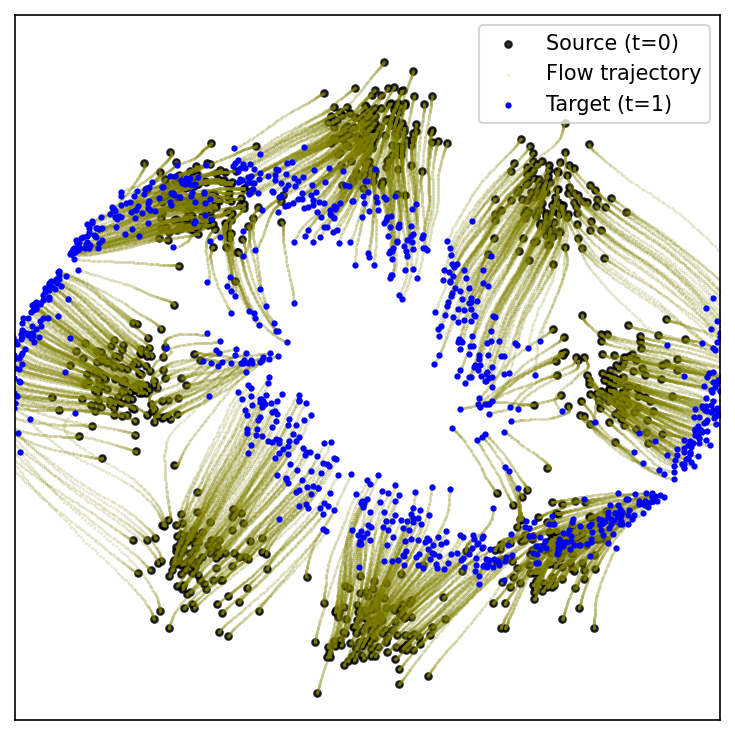}\hspace{2pt}
  \includegraphics[width=0.3\linewidth]{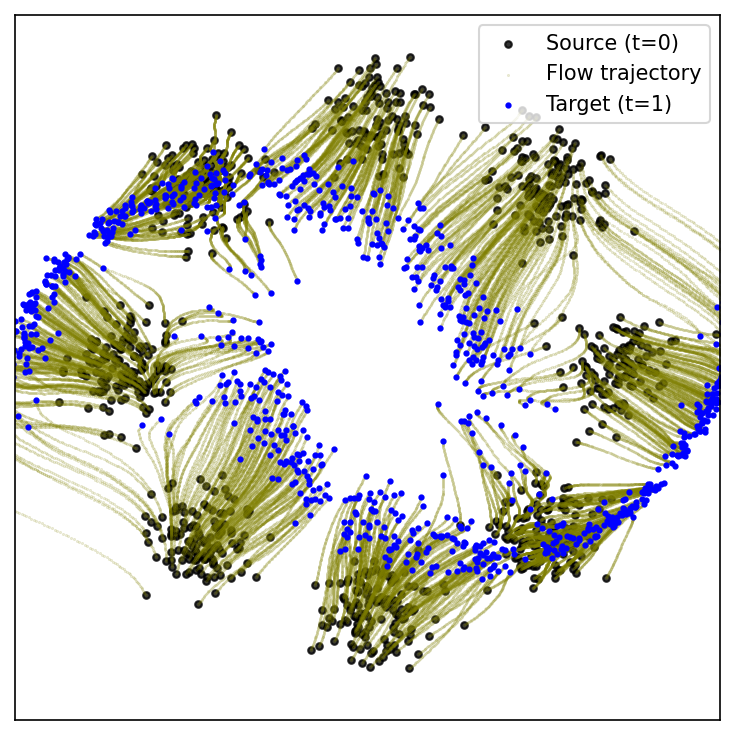}
  \\[2pt]
  % Bottom row: Gaussian → Checkerboard
  \includegraphics[width=0.3\linewidth]{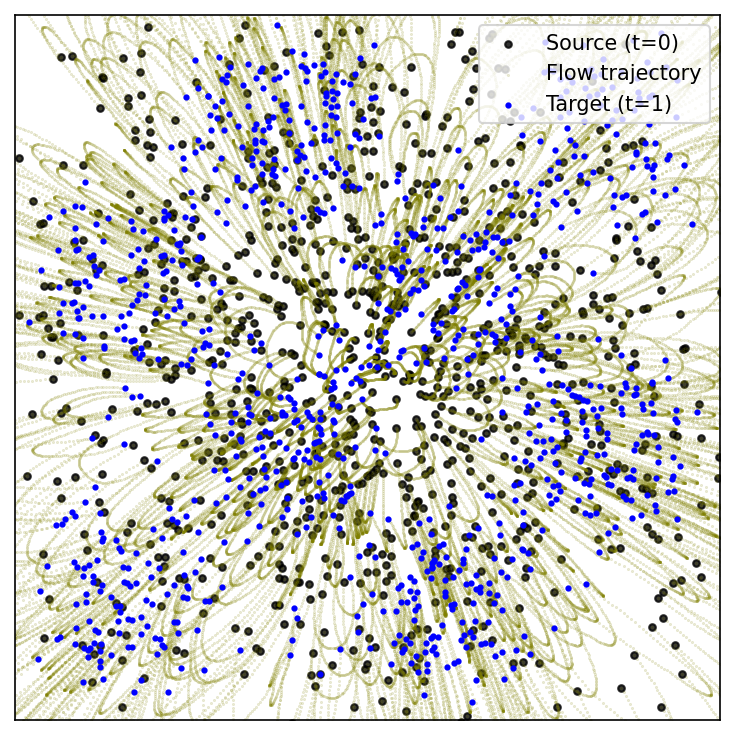}\hspace{2pt}
  \includegraphics[width=0.3\linewidth]{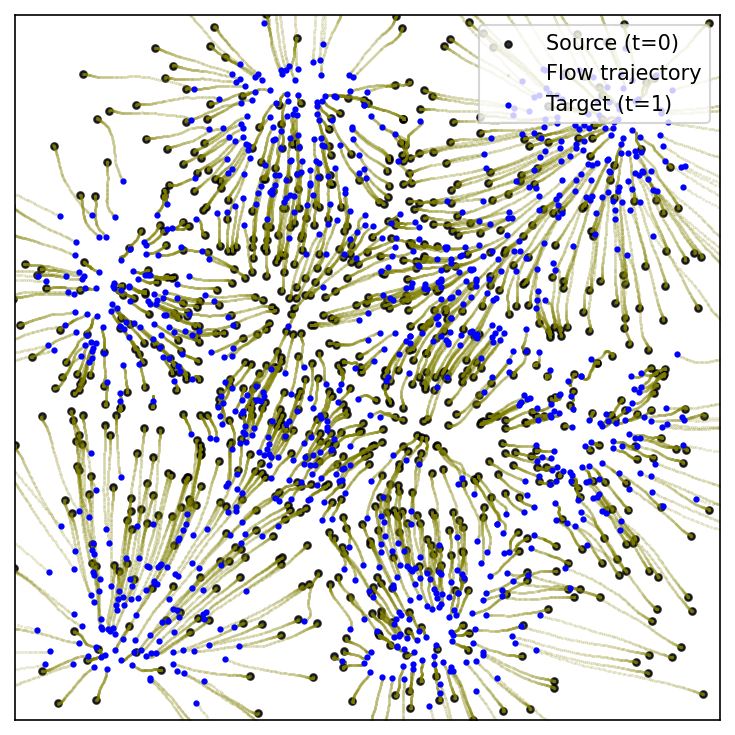}\hspace{2pt}
  \includegraphics[width=0.3\linewidth]{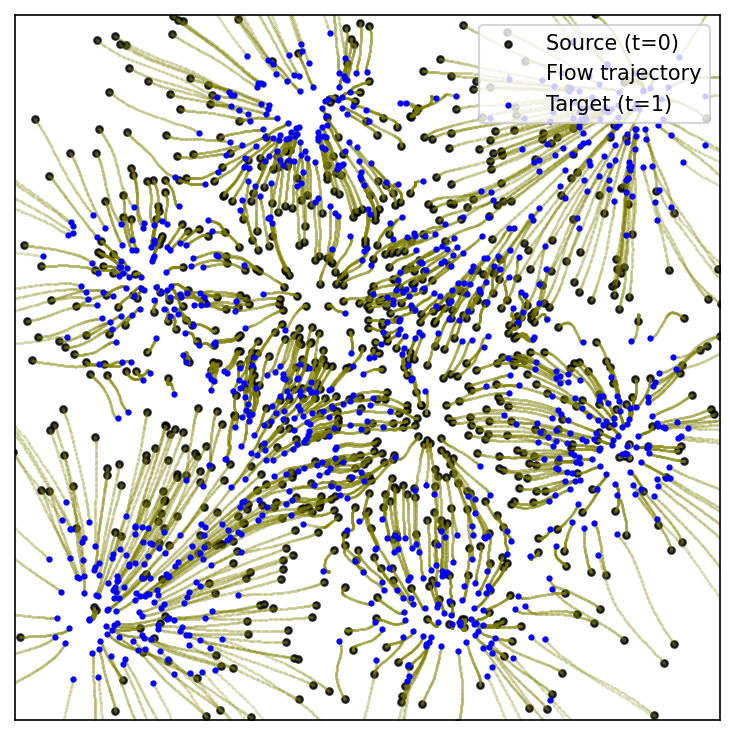}
  \caption{%
    \textbf{Trajectory ablation (global OT).}
    Top row: 8-GMM~$\to$~2-moons.
    Bottom row: Gaussian~$\to$~Checkerboard.
    Columns show Cosine (\textit{left}), Polynomial $\alpha{=}2$
    (\textit{center}), and Stochastic $\sigma{=}0.5$ (\textit{right}).
    Linear interpolation (Figure~\ref{fig:synthetic}) produces the
    straightest trajectories and is used throughout all other experiments.
    The disconnected checkerboard support amplifies trajectory failures,
    confirming linear interpolation as the optimal choice for
    neural flow matching.
  }
  \label{fig:traj_ablation}
\end{figure}

Figure~\ref{fig:traj_ablation} shows results on 8-GMM~$\to$~2-moons (top row)
and Gaussian~$\to$~Checkerboard (bottom row).
\textbf{Cosine} trajectories produce wildly curved paths that loop
outward before converging, requiring $g_\theta$ to model large
time-varying velocity changes that add unnecessary learning burden.
On the harder checkerboard benchmark, cosine trajectories collapse entirely,
failing to route mass into the disconnected squares.
\textbf{Polynomial} ($\alpha{=}2$) trajectories are cleaner but still
exhibit a slow start followed by rapid convergence near $t{=}1$,
introducing non-constant velocity that complicates learning and scatters
mass across square boundaries.
\textbf{Stochastic} ($\sigma{=}0.5$) interpolants inject time-dependent
noise that regularizes against mean collapse in the absence of OT, but
introduce endpoint variance that degrades one-step generation quality since
$F_\theta(1, \mathbf{x}_0)$ no longer targets $\mathbf{x}_1$ exactly.
In contrast, linear interpolation produces straight, consistent
trajectories with constant velocity $\dot{\mathbf{x}}_t = \mathbf{x}_1 - \mathbf{x}_0$,
minimizing the temporal modeling burden on $g_\theta$ and yielding
the sharpest one-step generation across both benchmarks.

\section{Conclusion}
\label{sec:conclusion}
We introduced Optimal Transport Neural Flow Matching (OT-NFM), an ODE-free
framework for one-step generative modeling that learns the flow map directly
using neural flows. By avoiding velocity-field parameterization and numerical
integration, OT-NFM reduces generation to a single forward pass. We identified \emph{mean collapse}, a failure mode unique to neural flow models
arising from inconsistent noise–data pairings, and showed theoretically that
optimal transport coupling is a necessary condition for non-degenerate flow map
learning. To make this practical for large-scale datasets, we proposed scalable
coupling strategies based on precomputed minibatch OT and online LOOM
refinement, both operating at $O(B^3)$ cost per batch. Experiments on synthetic transport benchmarks and image generation tasks
(MNIST and CIFAR-10) validate the theory: without OT, neural flows collapse to
the data mean, whereas consistent OT couplings produce sharp and diverse
samples. OT-NFM achieves competitive generation quality while requiring only a
single network evaluation at inference.

These results suggest that directly learning transport maps, combined with
structured coupling via optimal transport, provides a promising direction
toward efficient real-time generative modeling.
\bibliographystyle{splncs04}
\bibliography{main}
\end{document}